\documentclass[11pt,a4paper]{article}
\usepackage[T1]{fontenc}
\usepackage[utf8]{inputenc}
\usepackage{lmodern}
\usepackage{graphicx}
\usepackage{longtable}
\usepackage{tabularx}
\usepackage{array}
\usepackage{amsmath,amssymb}
\usepackage{float}
\usepackage{xcolor}
\usepackage{pifont}
\usepackage[hidelinks]{hyperref}
\title{Clinical radar heart sounds with echocardiographic phenotypes of ventricular dysfunction and valvular disease}
\author{Jinye Li, Qingchao Chen~\textsuperscript{\ding{41}}
\thanks{This work was supported by grants from the National Natural Science Foundation of China (62571009).}
\thanks{Jinye Li is with the National Institute of Health Data Science, the Institute of Medical Technology, Peking University. He is also with the Beijing University of Posts and Telecommunications. Qingchao Chen is with the National Institute of Health Data Science, the Institute of Medical Technology, and the State Key Laboratory of General Artificial Intelligence, Peking University. Email: qingchao.chen@pku.edu.cn.}
\thanks{Corresponding author: Qingchao Chen.}}
\date{}

\begin{document}
\maketitle
\begin{abstract}
Radar heart sounds capture cardiac mechanical activity without contact, while echocardiography characterizes ventricular and valvular function. Paired measurements connect these signals to structural heart disease, yet such data remain sparsely represented in public resources. Here we present 537 radar heart-sound recordings paired with one echocardiographic report for each of 162 patients. The cohort covers three overlapping phenotypes: left ventricular ejection fraction (LVEF) below 50\%, regional wall-motion abnormality and at least moderate valvular regurgitation, present in 38, 43 and 39 patients, respectively. Original complex signals, processed radar heart sounds and temporal quality segmentation support signal-to-phenotype analysis. Technical validation retains 95.93\% of signal duration and confirms reproducible waveform reconstruction. Patient-level five-fold evaluation achieves a mean outer-test AUROC of $0.770\pm0.138$ for reduced LVEF, identifying mechanical information associated with ventricular systolic dysfunction. This resource supports research linking non-contact cardiac mechanical signals to echocardiographic structural and functional phenotypes.
\end{abstract}
\section{Background \& Summary}
Structural heart disease, including ventricular dysfunction and valvular lesions, contributes to heart failure, disability and premature mortality, creating a substantial public-health burden\cite{roth2020burden,nkomo2006valves}. A US population-based study estimated moderate or severe valve disease in 2.5\% of adults and 13.3\% of those aged 75 years or older\cite{nkomo2006valves}. Echocardiography is a first-line imaging method and a clinical reference standard for characterizing cardiac structure and function. It quantifies left ventricular ejection fraction (LVEF), identifies regional wall-motion abnormalities and grades valvular regurgitation\cite{lang2015chambers,zoghbi2017valves}, providing clinically defined phenotypes of ventricular and valvular dysfunction.

Public radar resources have advanced non-contact assessment of cardiac timing and rhythm through recordings paired with ECG and other physiological references\cite{shi2020heartsounds,schellenberger2020radar,wu2026radar}. Representative collections include radar heart sounds with digital-stethoscope recordings, hemodynamic responses measured with impedance cardiography and blood pressure, and millimeter-wave signals referenced to ECG, respiration and pulse waveforms. These collections were acquired from healthy volunteers and do not pair radar recordings with echocardiographic labels of ventricular dysfunction or valve disease. Thus, the missing link is between radar-recorded cardiac mechanical activity and clinically characterized structural abnormalities. Addressing this gap requires patient recordings paired with measurements of systolic function, regional contraction and valvular regurgitation.

Here we present 537 radar heart-sound recordings paired with one retained echocardiographic report for each of 162 patients assessed in the Department of Cardiology at Peking University First Hospital. The cohort covers three nonexclusive phenotypes: LVEF below 50\%, regional wall-motion abnormality and at least moderate regurgitation of any valve, present in 38, 43 and 39 patients, respectively. Repeated recordings within patients capture cardiac mechanical waveforms, while original complex signals, processed radar heart sounds (RHS) and temporal quality segmentation support analyses at both signal and patient levels.

The paired resource supports investigation of how cardiac mechanical waveforms relate to ventricular systolic function, regional wall motion and valvular abnormalities. Technical validation retains 95.93\% of nominal signal duration and confirms reconstruction of processed RHS from original complex samples. Patient-level five-fold evaluation achieves a mean outer-test AUROC of $0.770\pm0.138$ for reduced LVEF, demonstrating identifiable mechanical information associated with systolic dysfunction. Together, clinical phenotyping, waveform traceability and patient-level evaluation provide a foundation for developing non-contact cardiac mechanical markers and investigating overlapping structural and functional abnormalities.

\section{Methods}
\subsection{Participants and clinical assessment}
The study was conducted in the Department of Cardiology at Peking University First Hospital among patients undergoing clinical cardiac assessment. Each patient underwent repeated one-minute radar measurements while resting supine. Measurements across visits were grouped within the same individual. The clinical cohort comprises 162 patients with numerical LVEF and binary classifications of regional wall-motion abnormality and valvular disease. Each patient is paired with one latest echocardiographic assessment, yielding 162 clinical assessments associated with 537 radar recordings.

Patients range from 18 to 89 years of age, with a mean of 63.22 years (SD 13.00); 109 are male and 53 female. LVEF is available for all 162 patients and ranges from 15.0\% to 80.1\%. Table~\ref{tab:cohort} summarizes the demographic and functional characteristics of the cohort and its overlapping phenotype-positive groups.

\begin{table}[htbp]
\centering\footnotesize
\caption{Demographic and LVEF distribution in the clinical cohort and phenotype-positive subgroups. Percentages use the column-specific denominator. Phenotype groups are nonexclusive. Each patient contributes one echocardiographic assessment. SD is the sample standard deviation; IQR denotes the first and third quartiles.}
\label{tab:cohort}
\begin{tabularx}{\textwidth}{>{\raggedright\arraybackslash}Xrrrr}
\hline
Characteristic & All & LVEF $<50\%$ & RWMA & Valve disease \\
\hline
Patients & 162 & 38 & 43 & 39 \\
Male & 109 (67.3\%) & 30 (78.9\%) & 29 (67.4\%) & 23 (59.0\%) \\
Female & 53 (32.7\%) & 8 (21.1\%) & 14 (32.6\%) & 16 (41.0\%) \\
Age, mean (SD), years & 63.22 (13.00) & 62.21 (16.43) & 64.19 (11.11) & 68.08 (11.69) \\
LVEF, mean (SD), \% & 59.75 (14.58) & 37.65 (8.68) & 50.64 (8.80) & 54.87 (17.96) \\
Age, median (IQR), years & 66.5 (56.2--72.0) & 68.0 (50.2--73.0) & 67.0 (56.5--73.0) & 69.0 (65.0--75.5) \\
LVEF, median (IQR), \% & 63.7 (50.8--71.2) & 38.7 (33.5--44.6) & 51.2 (46.7--57.5) & 59.5 (41.0--69.1) \\
Age 18--39 years & 7 (4.3\%) & 3 (7.9\%) & 0 (0.0\%) & 0 (0.0\%) \\
Age 40--49 years & 19 (11.7\%) & 6 (15.8\%) & 8 (18.6\%) & 4 (10.3\%) \\
Age 50--59 years & 25 (15.4\%) & 6 (15.8\%) & 5 (11.6\%) & 3 (7.7\%) \\
Age 60--69 years & 54 (33.3\%) & 6 (15.8\%) & 12 (27.9\%) & 13 (33.3\%) \\
Age 70--79 years & 48 (29.6\%) & 12 (31.6\%) & 17 (39.5\%) & 14 (35.9\%) \\
Age 80--98 years & 9 (5.6\%) & 5 (13.2\%) & 1 (2.3\%) & 5 (12.8\%) \\
LVEF $<$40\% & 20 (12.3\%) & 20 (52.6\%) & 5 (11.6\%) & 8 (20.5\%) \\
LVEF 40--$<$50\% & 18 (11.1\%) & 18 (47.4\%) & 12 (27.9\%) & 8 (20.5\%) \\
LVEF 50--$<$60\% & 26 (16.0\%) & 0 (0.0\%) & 20 (46.5\%) & 4 (10.3\%) \\
LVEF 60--$<$70\% & 52 (32.1\%) & 0 (0.0\%) & 6 (14.0\%) & 10 (25.6\%) \\
LVEF $\geq$70\% & 46 (28.4\%) & 0 (0.0\%) & 0 (0.0\%) & 9 (23.1\%) \\
\hline
\end{tabularx}
\end{table}

\subsection{Radar system and acquisition protocol}
\begin{figure}[htbp]
\centering
\includegraphics[width=\textwidth]{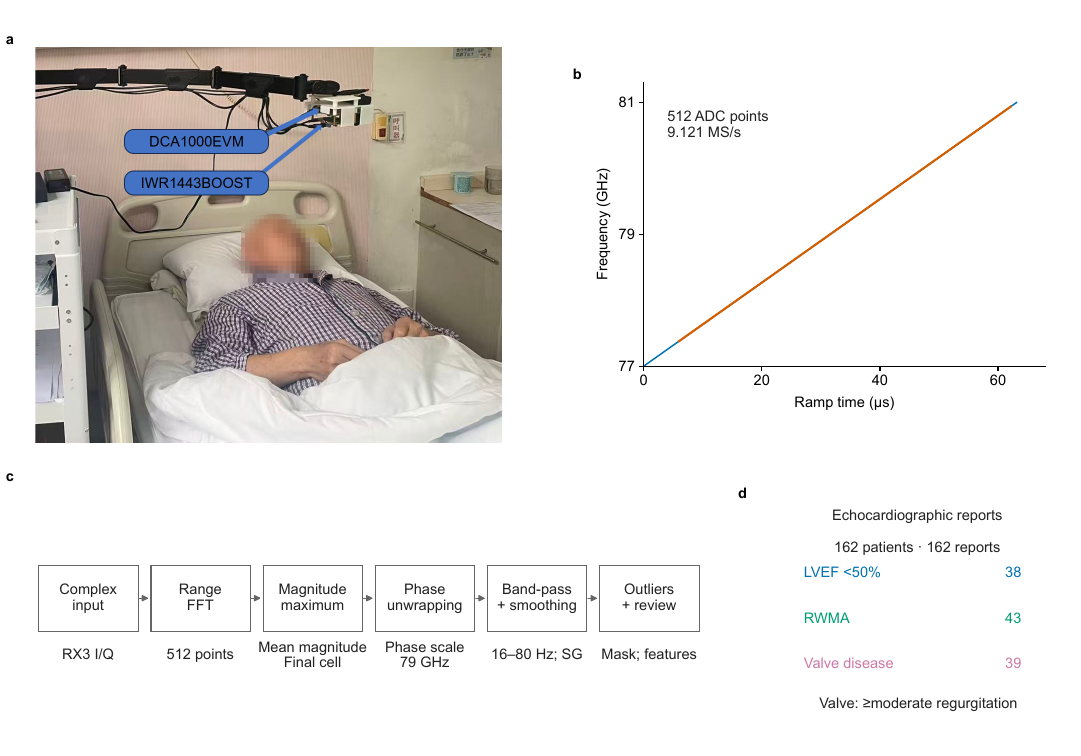}
\caption{Clinical acquisition environment, FMCW configuration and radar heart-sound workflow. a, Photograph of the clinical acquisition setup, showing a supine patient and the arm-mounted IWR1443BOOST radar and DCA1000EVM capture board positioned above the chest. b, FMCW ramp and sampled aperture: 77-GHz start, 63.343-MHz/$\mu$s slope, 63.14-$\mu$s ramp, 512 ADC samples at 9.121 MS/s, and one chirp per 1-ms frame. c, Complex-signal input, range FFT, maximum-magnitude range-cell selection, final-cell phase extraction, filtering and smoothing, followed by outlier removal and manual quality review. Offline reconstruction uses the final selected cell, preserving historical selection decisions. d, One retained report per selected patient supplies clinical context for three overlapping targets: LVEF below 50\%, regional wall-motion abnormality and at least moderate regurgitation of any valve.}
\label{fig:clinical-design}
\end{figure}

The radar system combined a Texas Instruments IWR1443 radio-frequency board with a DCA1000EVM acquisition board. One transmitting antenna and four receiving antennas were enabled (1TX4RX): TX0 was active, and TX1 and TX2 were disabled. Each chirp produced one slow-time observation. The processing path evaluated here uses the third receiving channel (RX3); its complex I/Q samples provide the input to heart-sound extraction. Figure~\ref{fig:clinical-design} shows the clinical acquisition environment and patient--radar placement alongside the FMCW configuration and analysis workflow.

The FMCW profile started at 77.000 GHz with a frequency slope of 63.343 MHz/$\mu$s. The ramp ended at 63.14 $\mu$s, giving the configured full-ramp bandwidth of approximately 3,999.48 MHz. Idle time was 10.00 $\mu$s, TX start time 1.00 $\mu$s and ADC start time 6.00 $\mu$s. Each chirp contained 512 complex ADC samples per receive channel at 9.121 MS/s. The sampled aperture therefore lasted approximately 56.13 $\mu$s and occupied approximately 3.556 GHz of the ramp. The nominal range resolution $c/(2B)$ is approximately 3.75 cm when calculated from the full ramp, or 4.22 cm using the sampled aperture. The receive gain was 30 dB, and the first and second analog high-pass corner settings were 175 and 350 kHz, respectively.

Each frame contained one chirp, with a frame period of 1 ms. A recording consisted of 60,000 frames, corresponding to a slow-time sampling rate of 1,000 Hz and a duration of 60 s. Acquisition was initiated by software triggering with zero frame-trigger delay. The 9.121-MS/s fast-time rate resolves the sampled chirp, whereas the 1,000-Hz slow-time rate tracks the phase evolution of cardiac-related vibration.

A complex radar acquisition is represented by slow-time observations and fast-time samples. For the evaluated receive channel, each recording has 60,000 observations with 512 complex samples per observation. Heart-sound extraction transforms this two-dimensional signal into a 60,000-sample vibration time series. Table~\ref{tab:acquisition} summarizes the acquisition parameters and the settings used for waveform and spectral analysis.

Patients rested supine during recordings of cardiac-related chest-surface vibration. The radar faced the anterior chest and was aimed approximately 2 cm inferolateral, towards the patient's left, to the intersection of the chest midline and the line joining the nipples. The radar--chest distance was approximately 50--80 cm. This landmark served as the reference for repeated one-minute acquisitions over anterior chest regions. Clothing was unrestricted, and the protocol used resting measurements. Automatic target-cell selection used the largest time-averaged range-profile magnitude. The collection comprises radar recordings paired with echocardiographic reports; ECG was not acquired.

\begin{table}[htbp]
\centering\footnotesize
\caption{TI IWR1443/DCA1000EVM acquisition configuration and offline analysis settings. Full-ramp bandwidth and sampled bandwidth are distinct; the 1-ms frame period defines the exported 1,000-Hz vibration time series.}
\label{tab:acquisition}
\begin{tabularx}{\textwidth}{>{\raggedright\arraybackslash}p{0.31\textwidth} >{\raggedright\arraybackslash}p{0.26\textwidth} >{\raggedright\arraybackslash}X}
\hline
Parameter & Setting & Interpretation \\
\hline
RF board / acquisition board & TI IWR1443 / DCA1000EVM & Non-contact FMCW front end and raw-data capture \\
Transmit / receive configuration & 1TX4RX & TX0 enabled; TX1/TX2 disabled; four complex receive channels \\
Chirp start frequency & 77.000 GHz & FMCW profile start \\
Frequency slope & 63.343 MHz/$\mu$s & Linear frequency modulation \\
Idle / TX start time & 10.00 / 1.00 $\mu$s & Chirp timing settings \\
ADC start / ramp-end time & 6.00 / 63.14 $\mu$s & Sampled aperture within the ramp \\
ADC samples / fast-time rate & 512 / 9.121 MS/s & Per chirp and per receive channel \\
Full-ramp / sampled bandwidth & 3,999.48 / approx. 3,555.64 MHz & Full profile versus 56.13-$\mu$s sampled aperture \\
Nominal range resolution & 3.75 / 4.22 cm & Full-ramp / sampled-aperture calculation \\
Receive gain / HPF corners & 30 dB / 175 and 350 kHz & Receiver analog settings \\
Chirp / loop configuration & Index 0; one loop per frame & No transmitter multiplexing in the exported recording \\
Frame period / number of frames & 1 ms / 60,000 & Recording timing \\
Trigger / trigger delay & Software / 0 $\mu$s & Frame acquisition start \\
Slow-time rate / duration & 1,000 Hz / 60 s & One temporal observation per frame \\
ADC quantization & 16-bit per I/Q component & Digitized complex I/Q measurements \\
RHS length / reconstruction channel & 60,000 / RX3, one-based & Continuous waveform; explicit single-channel reconstruction \\
Target-cell selection & Maximum mean magnitude & Peak within the configured search region; reconstruction uses the final selected cell \\
Acquisition posture / state & Supine; resting & Repeated chest acquisitions; no specific clothing requirement \\
Chest reference / radar distance & Approximately 2 cm inferolateral / 50--80 cm & Patient's left from the midline--nipple-line intersection \\
Clinical correspondence & One latest report per patient & Every included radar recording is within three calendar months \\
Quality-monitoring sequence & Outlier removal, then manual inspection & Final reviewed interval mask used for analysis \\
Phase-amplitude conversion constant & 79 GHz & Phase-amplitude conversion \\
RHS band-pass / smoothing & Order 4, 16--80 Hz / 31 points, degree 3 & Forward--backward Butterworth and Savitzky--Golay \\
Envelope / envelope smoothing & Hilbert magnitude / 101 points, degree 3 & Illustrative signal representation \\
Spectral window lengths / overlap & 2,048 and 4,096 points / 50\% & Fully retained windows; edge exclusions \\
\hline
\end{tabularx}
\end{table}

\subsection{Extraction of radar heart sounds}
Let $s[n,m]$ denote the complex I/Q signal for the evaluated receive channel, with slow-time index $n$ and fast-time index $m$. The fast-time mean is removed independently for each chirp, and a 512-point Fourier transform yields the complex range profile $S[n,k]$. The target cell maximizes the slow-time mean magnitude within the candidate region: $A[k]=N^{-1}\sum_{n=1}^{N}|S[n,k]|$ and $k^*=\arg\max_{k\in\mathcal{R}} A[k]$, where $\mathcal{R}$ denotes the candidate cells. One cell is fixed for phase tracking throughout a recording. Historical manual adjustment of the automatically selected cell occurred in 60 of the 537 signal-validation recordings; waveform reconstruction uses the final selected cell in each case.

For the selected range-cell trace $z[n]$, the unwrapped phase is $\phi[n]=\operatorname{unwrap}(\arg z[n])$. The amplitude convention used during processing is $a[n]=(c/f_{\mathrm{conv}})\phi[n]/(2\pi)$, with $f_{\mathrm{conv}}=79$ GHz. This conversion constant is retained separately from the 77-GHz start frequency of the acquisition profile. Waveform illustrations and reconstruction comparisons use normalized amplitude, with each compared signal independently mean-centred and RMS-normalized.

A fourth-order Butterworth band-pass design at 16--80 Hz is applied forward and backward to obtain a zero-net-phase-delay offline waveform. The filtered signal is smoothed with a 31-point, third-degree Savitzky--Golay filter\cite{savitzky1964smoothing}. The resulting signed mechanical vibration series is the radar heart sound (RHS). Its Hilbert analytic-signal magnitude supplies an amplitude envelope, which is smoothed with a 101-point, third-degree Savitzky--Golay filter for illustration. The signed waveform and envelope provide separate signal representations.

Signal-quality monitoring combines automated outlier removal with subsequent manual inspection. The quality-review implementation identifies large deviations from the median using a median-absolute-deviation-based robust scale, with a standard-deviation fallback when that scale is zero. Flagged isolated samples are interpolated on the original sampling grid for inspection. Reviewers then inspect the signals and mark excluded time intervals. Validation applies the reviewed interval mask to the RHS. Exclusion intervals are unioned, and the original time axis is maintained for all retained samples.

\subsection{Echocardiographic measurements and phenotype definitions}
Echocardiographic assessment characterizes chamber dimensions, ventricular volumes, LVEF, regional contraction, mitral inflow, tissue Doppler and valvular function. The latest examination is retained for each patient. LVEF measurements include Teichholz and biplane Simpson approaches, alongside two-dimensional/M-mode and Doppler assessment. The analysis uses the reported clinical measurements.

Reduced LVEF is defined by the strict threshold $\mathrm{LVEF}<50\%$, applied to the reported clinical LVEF. This phenotype is present in 38 patients, with 124 above or at the threshold. Regional wall-motion abnormality is represented as a binary clinical phenotype and is present in 43 patients. These targets characterize global systolic function and regional contraction, respectively.

Valvular disease is defined as at least moderate regurgitation of any valve, including the aortic, mitral, tricuspid or another valve. It is present in 39 patients, with 123 negative for this target. Mild regurgitation alone and stenosis alone are negative under this definition. The three phenotypes are nonexclusive, reflecting the coexistence of ventricular and valvular abnormalities. Lesion-specific echocardiographic evaluation of regurgitation is described in clinical recommendations\cite{zoghbi2017valves}.

\subsection{Pairing of radar and echocardiographic measurements}
All repeated radar measurements from one patient are paired with that patient's latest echocardiographic assessment. Patient-level grouping is maintained across visits, so that each individual contributes one clinical phenotype vector and multiple radar observations. The paired design contains 162 clinical assessments and 537 radar recordings.

The pairing protocol specifies an interval of no more than three calendar months between radar measurement and echocardiographic examination, in either direction. Clinical summaries use one observation per patient, whereas signal-level experiments use individual radar recordings.

\subsection{Offline validation experiments and statistical analysis}
Signal-level technical validation examines 537 recordings to assess continuity, retained exposure, spectral sensitivity and processing reproducibility. Numerical checks evaluate RHS length, finite values and nonconstant content. Reviewed exclusion intervals are unioned and applied as a temporal mask on the original time axis. Retained duration is calculated before excluding the first and last second for spectral analysis. These edges are omitted uniformly to reduce edge influence. Excluded intervals include their start sample and omit their end sample.

Welch-style spectra\cite{welch1967spectra} average periodic-Hann-windowed, constant-detrended periodograms of 2,048-sample windows with 50\% overlap. Only windows entirely within retained data contribute. The 16--80-Hz centroid is $C=\sum fP(f)/\sum P(f)$, with both sums restricted to that band. A band-power fraction uses 16--80\,Hz in the numerator and 1--200\,Hz in the denominator. This fraction quantifies the spectral concentration of the filtered and smoothed RHS representation.

Within-record variation is evaluated by comparing centroids in the first and last 30\,s under the same quality mask. Window-length sensitivity is evaluated with 2,048- and 4,096-sample windows. Absolute differences and descriptive Spearman correlation quantify these comparisons. Processing reproducibility is evaluated in 12 deterministically selected recordings. The complex RX3 signal undergoes fast-time centring, range transformation, final-cell phase extraction and the specified filtering and smoothing. RHS and reconstructed waveforms are independently mean-centred and RMS-normalized after removing one second from each edge. Pearson correlation and normalized RMSE assess reproduction of normalized waveform shape.

Clinical summaries use one observation per patient and available finite values independently per variable. Standard deviations use the sample denominator, and quartiles use linear interpolation. Values are analyzed without imputation or winsorization. Binary targets retain valid zero-or-one entries, with other values classified as invalid or missing. The three disease axes are nonexclusive, and each numerical field has its own available-value denominator.

\subsection{Patient-level evaluation of reduced LVEF}
To evaluate whether radar cardiac mechanical signals characterize left ventricular systolic function, an LVEF classification experiment used echocardiographic LVEF as the clinical reference and defined $\mathrm{LVEF}<50\%$ as the positive target. The experiment evaluated all 162 patients in the study cohort: 124 with LVEF at or above 50\% and 38 below 50\%.

The RHS waveform was resampled to 500 Hz and divided into nonoverlapping 5-s windows. Within each outer fold, a global median and interquartile range were estimated exclusively from the training set and used for robust scaling of the training, validation and test signals. Applying the same training-derived scale across patients preserves their relative amplitude differences. The model jointly encoded the RHS time-domain waveform, Hilbert envelope and logarithmic short-time Fourier transform (STFT) representation. Recording-level multiple-instance learning (MIL) aggregated consecutive windows within each recording. Training used focal loss with automatically determined class weights to accommodate the lower prevalence of the positive target.

Evaluation used five outer cross-validation folds grouped by patient identity. In each fold, approximately 20\% of patients formed the outer test set, while the remaining patients were divided into training and validation sets corresponding to approximately 70\% and 10\% of the full cohort, respectively. All recordings and windows from the same patient remained in one partition. The validation set was used for model selection, and each outer test set was used for final evaluation. Performance was assessed at the patient level using the area under the receiver operating characteristic curve (AUROC), area under the precision--recall curve (PR-AUC), macro-averaged F1 and accuracy.

\subsection{Distribution and overlap of clinical phenotypes}
\begin{figure}[htbp]
\centering
\includegraphics[width=\textwidth]{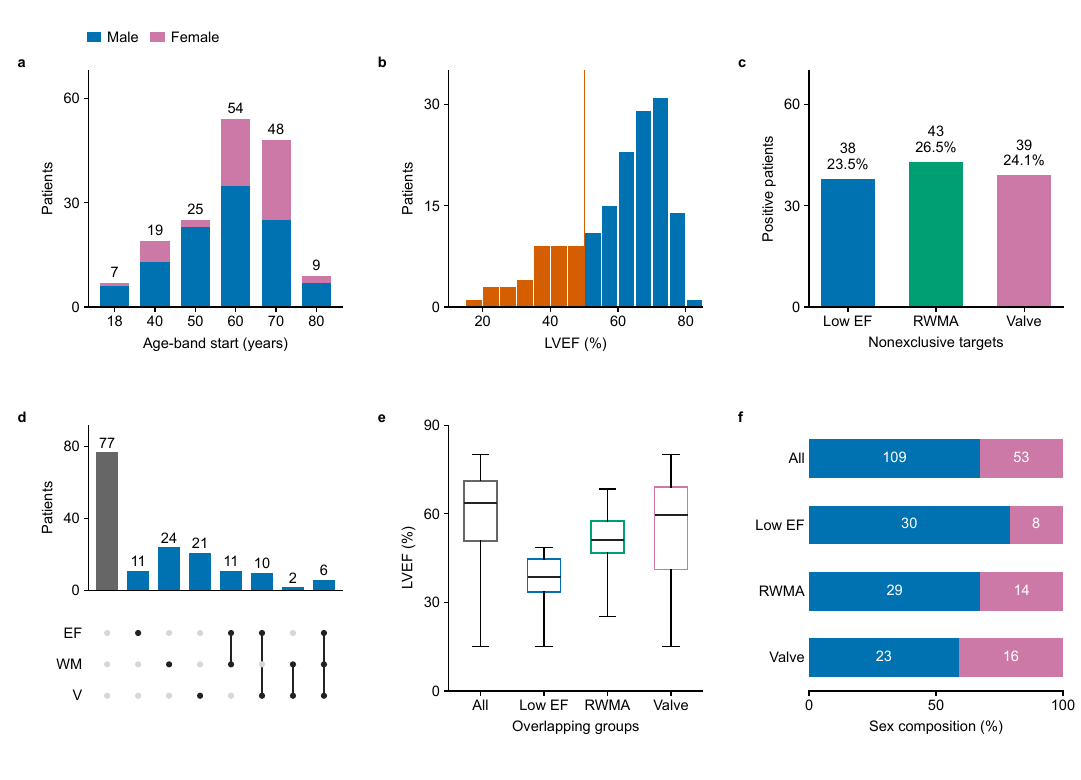}
\caption{Demographic distribution and overlapping clinical phenotypes in 162 patients, each with one retained report. a, Age-band counts stacked by recorded sex; axis labels give the lower age bound. b, Reported clinical LVEF in 5-percentage-point bins, with the orange line denoting 50\% and orange bars denoting values strictly below it. c, Positive counts and percentages for the three nonexclusive targets. Valvular disease means at least moderate regurgitation of any valve. d, Joint label counts; filled-dot rows EF, WM and V denote reduced LVEF, regional wall-motion abnormality and valvular disease. The empty intersection denotes the group with none of the three target labels. e, LVEF by nonexclusive group; boxes span the first to third quartile, horizontal lines show medians, and whiskers span the observed minimum to maximum. f, Sex composition of the whole cohort and phenotype-positive groups, with counts inside the bars. Clinical distributions use one observation per patient.}
\label{fig:phenotypes}
\end{figure}

The cohort contains 38 patients with LVEF below 50\% (23.46\%), 43 with regional wall-motion abnormalities (26.54\%) and 39 with valvular disease (24.07\%). All percentages use the same denominator of 162 patients, each with one retained report. Mean LVEF is 59.75\% (SD 14.58), with median 63.65\% and interquartile range 50.75--71.18\%. Figure~\ref{fig:phenotypes} shows the distribution and intersections of these labels.

The clinical population is predominantly middle-aged and older. The age bands 18--39, 40--49, 50--59, 60--69, 70--79 and 80 years or older contain seven, 19, 25, 54, 48 and nine patients, respectively; 111 of 162 (68.52\%) are aged at least 60 years. Male and female patients account for 67.28\% and 32.72\%. Numerical LVEF is below 40\% in 20 patients and between 40\% and strictly below 50\% in 18; the remaining 26, 52 and 46 patients lie in the 50--below-60\%, 60--below-70\% and at-least-70\% bands, respectively. Repeated radar acquisitions do not increase these patient-level denominators.

Table~\ref{tab:cohort} summarizes demographic and LVEF variation across the overlapping phenotype-positive subgroups. Mean age is 62.21 years (SD 16.43) in the reduced-LVEF group, 64.19 (11.11) in the wall-motion group and 68.08 (11.69) in the valve group. Mean LVEF is 37.65\%, 50.64\% and 54.87\%, respectively. These summaries describe each clinical axis within the same patient population.

Among the 38 patients with reduced LVEF, 17 also have wall-motion abnormality and 16 have valvular disease; six carry all three labels. Overall, 56 patients carry one target, 23 carry two and six carry all three. The remaining 77 form the target-negative group for these three phenotypes. This overlap motivates multilabel analysis of the clinical targets.

The echocardiographic assessments describe the same 162-patient population throughout the demographic summaries and phenotype plots. Each numerical variable uses the denominator of patients with an available measurement. Patient-level summaries preserve the distinction between clinical variation and repeated radar observations.

\section{Data Records}
The radar measurements are represented at three complementary levels: the original complex signal, the processed RHS and temporal quality segmentation (Table~\ref{tab:records}). These three components link the acquired I/Q signal to the vibration waveform and its analysis intervals. The original complex signal and processed RHS are provided in MATLAB format.

The complex signal retains the fast-time and slow-time dimensions used for range selection and phase tracking. The evaluated RX3 representation contains 60,000 slow-time observations and 512 complex fast-time samples per observation. This representation supports re-extraction of cardiac mechanical signals with alternative range-selection or filtering strategies.

Each RHS is a continuous 60,000-sample signed waveform at 1,000 Hz. Temporal quality segmentation marks retained and excluded intervals following outlier screening and manual review. These intervals remain aligned with the original 60-s time axis, enabling consistent calculation of retained exposure and quality-aware spectral features.

The measurements span the echocardiographic phenotypes described in the Methods. Repeated recordings share a patient's clinical phenotype vector, supporting both within-patient signal analysis and comparisons across the overlapping clinical groups.

\begin{table}[htbp]
\centering\footnotesize
\caption{Three complementary representations of each radar measurement. MATLAB format is used for the complex signal and processed RHS; quality segments retain their positions on the original time axis.}
\label{tab:records}
\begin{tabularx}{\textwidth}{>{\raggedright\arraybackslash}p{0.25\textwidth} >{\raggedright\arraybackslash}p{0.23\textwidth} >{\raggedright\arraybackslash}X}
\hline
Representation & Extent & Analytical role \\
\hline
Original complex radar signal & 60,000 by 512 per recording & Complex I/Q representation for the evaluated RX3 channel; supports range selection and phase-based signal extraction \\
Processed RHS & 60,000 samples at 1,000 Hz per recording & Signed cardiac-related vibration waveform; normalized amplitude is used for waveform illustrations and reconstruction comparisons \\
Temporal quality segmentation & Reviewed intervals on each 60-s time axis & Retained and excluded intervals following outlier screening and manual review; supports quality-aware exposure and spectral analysis \\
\hline
\end{tabularx}
\end{table}

\begin{figure}[htbp]
\centering
\includegraphics[width=\textwidth]{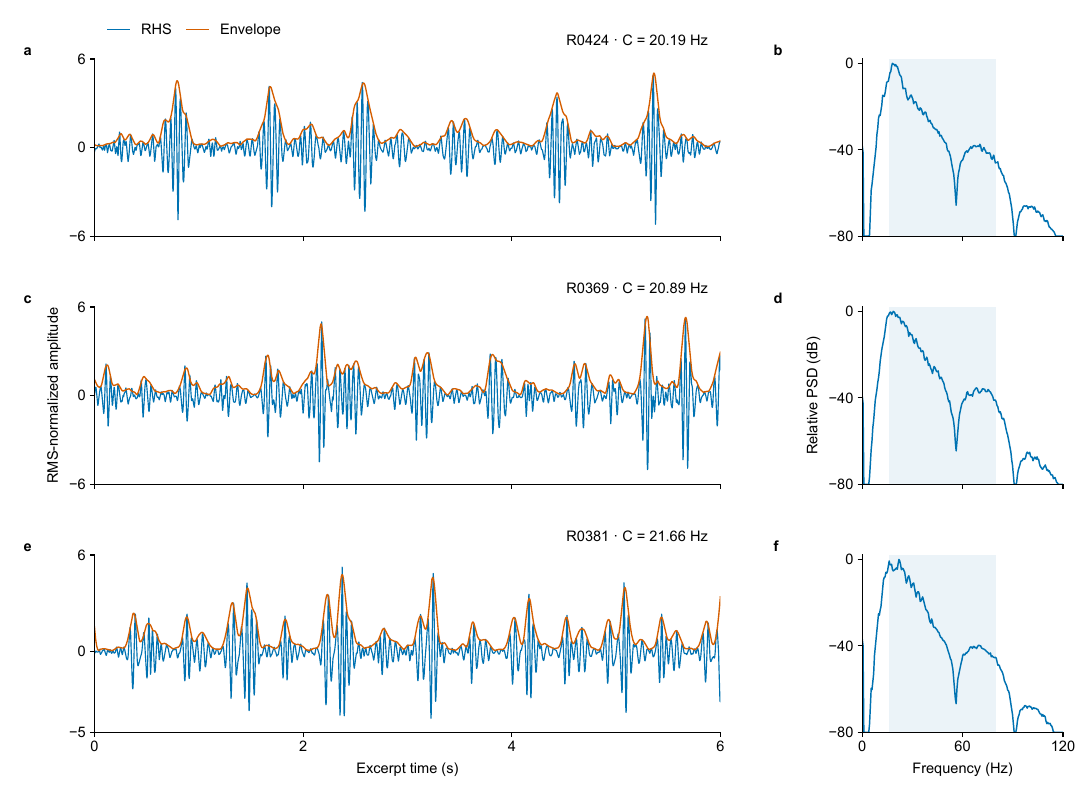}
\caption{Representative radar heart-sound waveforms and their spectra. a, c and e, Six-second excerpts from three recordings selected at approximately the first quartile, median and third quartile of the quality-masked spectral-centroid distribution. Blue denotes the signed RHS waveform and orange its smoothed Hilbert envelope; both use the RMS of the displayed signed excerpt as the amplitude scale. Excerpts lie within retained intervals. b, d and f, Quality-masked spectra from the respective full recordings, normalized to their maximum power; shading indicates 16--80 Hz. R-identifiers denote anonymous analytical recordings. The examples illustrate variation across the spectral-centroid distribution.}
\label{fig:rhs-examples}
\end{figure}

\section{Technical Validation}
\subsection{Signal continuity and quality-aware exposure}
All 537 RHS vectors in the signal-validation set contain 60,000 finite, nonconstant samples. Effective excluded intervals lie within the one-minute record. After interval union, kept and deleted durations sum to the nominal duration for all 537 recordings within a tolerance of 2 ms. These checks establish numerical continuity and consistent temporal accounting of the reviewed quality intervals.

\begin{figure}[htbp]
\centering
\includegraphics[width=\textwidth]{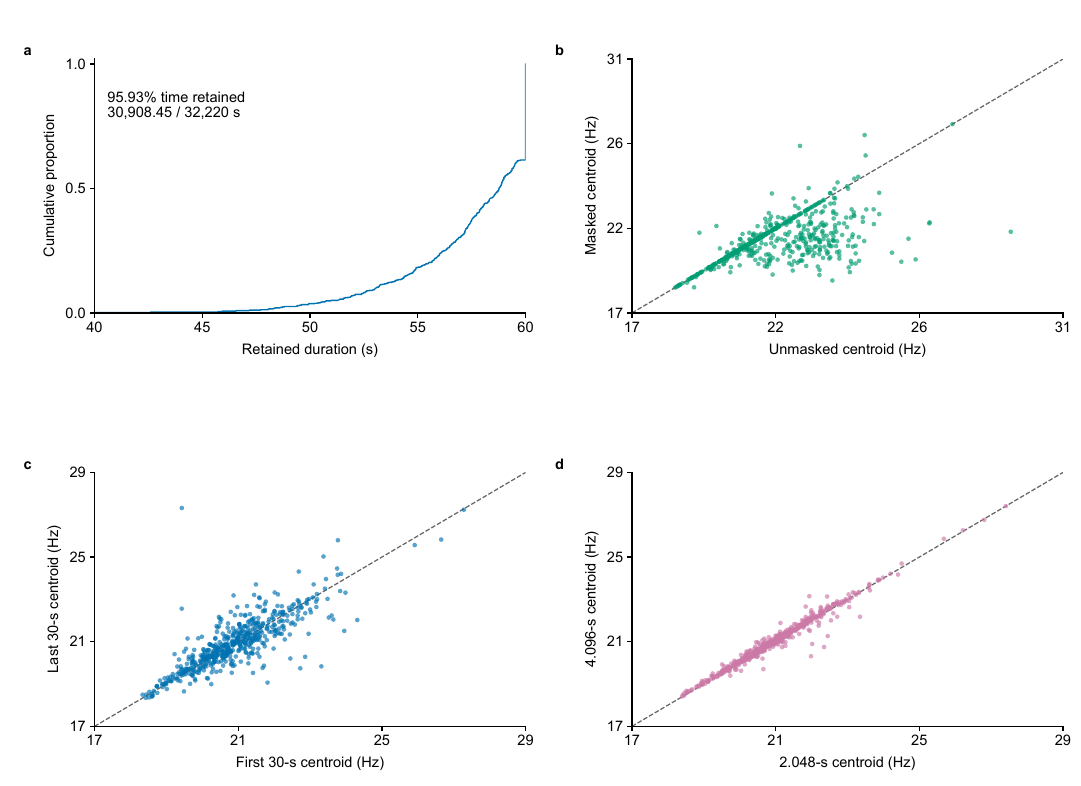}
\caption{Quality-aware signal validation and analysis sensitivity in 537 selected recordings. a, Empirical cumulative distribution of retained duration, calculated from the union of effective deleted intervals before record-edge exclusion. b, Centroid before and after quality masking, with the same one-second edge exclusions. c, Quality-masked centroids in the first and last 30 s. d, Quality-masked centroids using 2.048- and 4.096-s windows, each with 50\% overlap. Centroids are restricted to 16--80 Hz. Each point is one recording; gray dashed lines indicate equality. Comparisons use the final reviewed quality masks.}
\label{fig:signal-validation}
\end{figure}

Quality exclusions affect 330 recordings (61.45\%). The retained duration totals 30,908.45 s, or 8.59 h, corresponding to 95.93\% of the nominal 32,220 s. Median retained duration is 58.77 s (interquartile range 56.25--60.00), and the minimum is 40.04 s. The time excluded at the two record edges for spectral analysis is additional to these coverage estimates. Figure~\ref{fig:signal-validation}a shows the cumulative retained-duration distribution.

\subsection{Spectral sensitivity to quality masking and window length}
Every selected recording contributes at least one complete quality-retained window to both spectral window-length analyses and both recording-half analyses. The quality-masked 16--80-Hz centroid has a median of 20.89 Hz (interquartile range 20.19--21.66). The median band-power fraction is 0.873 (0.826--0.905). The spectral distribution describes the RHS representation after band-pass filtering and smoothing.

The median absolute centroid change after applying the quality mask is 0.213 Hz (interquartile range 0--1.534), with a maximum of 7.828 Hz. These differences demonstrate the sensitivity of energy-weighted features to temporal quality segmentation: excluding a small fraction of the duration can substantially change the spectral centroid.

Increasing the analysis window from 2.048 to 4.096 s changes the centroid by a median absolute 0.073 Hz (0.035--0.148), with a maximum of 1.671 Hz. The first and last 30-s centroids have descriptive Spearman correlation 0.815 and median absolute difference 0.279 Hz (0.138--0.600); the maximum half difference is 7.888 Hz. Window-length sensitivity is typically smaller than the change associated with temporal masking or within-record variation, but individual recordings can differ. These comparisons quantify analytical sensitivity and temporal variation within a saved recording.

\subsection{Reconstruction from raw complex radar samples}
\begin{figure}[htbp]
\centering
\includegraphics[width=\textwidth]{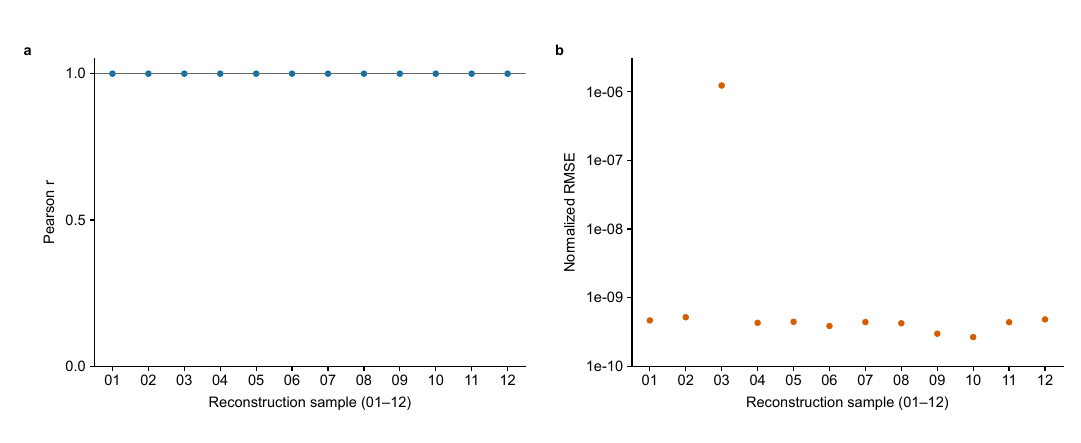}
\caption{Reconstruction of RHS from the original complex radar signal. a, Pearson correlation of RHS and reconstructed waveforms in 12 deterministically selected recordings. b, Corresponding RMSE after independent mean centring and RMS normalization; the RMSE axis is logarithmic. Both comparisons omit the first and last second. Numbers 01--12 denote the sampled recordings used for normalized waveform-shape reconstruction.}
\label{fig:reproduction}
\end{figure}

The raw-data experiment reconstructs 12 selected recordings using the final selected range cell and the evaluated single-channel processing path. All 12 RHS and reconstructed normalized waveforms have Pearson correlation above 0.999999. Normalized RMSE ranges from $2.65\times10^{-10}$ to $1.23\times10^{-6}$ (Fig.~\ref{fig:reproduction}). These results establish close agreement between the processed RHS and the waveform reconstructed from the original complex radar signal.

The reconstruction experiment evaluates decoding, final-cell phase extraction and filtering/smoothing in the 12 sampled records. Independent mean centring and RMS normalization make the comparison a test of waveform shape. Table~\ref{tab:validation} summarizes the executed experiments and their observation units.

\subsection{Clinical reference and patient-level phenotypes}
The paired design retains one latest echocardiographic assessment for each of 162 patients. Clinical LVEF and the strict 50\% threshold identify 38 positive and 124 negative patients. Binary classifications identify 43 patients with regional wall-motion abnormality and 39 with at least moderate valvular regurgitation. Patient-level grouping links repeated mechanical observations to the corresponding clinical assessment.

\subsection{Discrimination of reduced left ventricular systolic function}
Patient-level five-fold cross-validation in the 162-patient study cohort yielded a mean outer-test AUROC of $0.770\pm0.138$. The reported PR-AUC was 0.559, macro-averaged F1 was 0.652 and accuracy was 71.6\% (Table~\ref{tab:validation}). These results connect the radar-derived mechanical representation to an echocardiographically defined functional target: non-contact RHS contains information associated with reduced LVEF that can be identified at the patient level. This functional evaluation complements the signal-quality and reconstruction experiments, supporting the use of paired radar and echocardiographic measurements to investigate ventricular systolic dysfunction.

\begin{table}[htbp]
\centering\footnotesize
\caption{Signal and functional validation. Medians of signal metrics are followed by interquartile ranges. Signal-level observations use 537 recordings, including 12 sampled records for reconstruction; LVEF discrimination uses the 162-patient study cohort. AUROC summarizes the five outer test folds.}
\label{tab:validation}
\begin{tabularx}{\textwidth}{>{\raggedright\arraybackslash}p{0.28\textwidth} >{\raggedright\arraybackslash}p{0.17\textwidth} >{\raggedright\arraybackslash}X}
\hline
Experiment / metric & Unit / $n$ & Observed result \\
\hline
RHS numerical integrity & Recordings / 537 & All 60,000-point vectors finite and nonconstant \\
Quality interval accounting & Recordings / 537 & No kept-plus-deleted duration discrepancy above 2 ms \\
Retained signal exposure & Recordings / 537 & 30,908.45 s total; 95.93\% of nominal duration \\
Retained duration, s & Recordings / 537 & 58.77 (56.25--60.00); minimum 40.04 \\
Quality-masked centroid, Hz & Recordings / 537 & 20.89 (20.19--21.66) \\
Processed band-power fraction & Recordings / 537 & 0.873 (0.826--0.905) \\
Absolute masking change, Hz & Recordings / 537 & 0.213 (0--1.534); maximum 7.828 \\
Absolute half-record difference, Hz & Recordings / 537 & 0.279 (0.138--0.600); Spearman correlation 0.815 \\
Absolute window-length change, Hz & Recordings / 537 & 0.073 (0.035--0.148); maximum 1.671 \\
Raw-to-RHS normalized shape & Recordings / 12 & Pearson correlation above 0.999999 in every sampled record \\
Raw-to-RHS normalized RMSE & Recordings / 12 & $2.65\times10^{-10}$ to $1.23\times10^{-6}$ \\
\hline
Reduced-LVEF AUROC & Patients / 162 & $0.770\pm0.138$; mean across five outer test folds \\
Reduced-LVEF PR-AUC & Patients / 162 & 0.559 \\
Reduced-LVEF macro-F1 & Patients / 162 & 0.652 \\
Reduced-LVEF accuracy & Patients / 162 & 71.6\% \\
\hline
\end{tabularx}
\end{table}

\section{Usage Notes}
The paired recordings support cardiac-mechanical research with echocardiographic clinical context. Applications include quality-aware feature extraction, comparison of processing methods and multilabel analysis of ventricular dysfunction, regional contraction and valvular regurgitation. Raw complex samples permit alternative signal representations to be evaluated against the processed RHS.

Quality-aware analysis should preserve the original time axis and use windows wholly contained in retained intervals. A minimum retained-duration criterion should be specified before supervised modelling. The supplied masks and retained-duration summaries provide the temporal information needed to apply such a criterion.

Clinical analyses should preserve the nonexclusive clinical phenotype definitions. Reduced LVEF uses the reported clinical value and the strict 50\% threshold; valvular disease uses at least moderate regurgitation of any valve. The 77 patients with no target labels form the three-target-negative group. Report-level measurement methods, age, sex and coexisting phenotypes are relevant covariates for subsequent association analyses.

Training, validation and test partitions should be formed at patient level, with all recordings across visits from one patient assigned to the same partition. Subsequent modelling should report patient-level denominators, label overlap, missingness and class-specific performance with uncertainty. This grouping aligns model evaluation with the patient-level echocardiographic targets.

\section{Code Availability}
The study software implements radar control, complex-signal processing, range selection, phase-based RHS extraction, quality segmentation and the technical validation experiments. The evaluated operations and parameters are specified in the Methods. The acquisition and analysis software is maintained with the study materials.

\section{Data Availability}
The radar measurements are maintained by the investigators as original complex-signal MATLAB data, processed RHS MATLAB data and temporal quality segmentation. Each measurement is associated with the corresponding patient-level echocardiographic phenotype.

\bibliographystyle{unsrt}
\bibliography{references}
\IfFileExists{supplementary/supplementary.tex}{\clearpage\appendix\section{Supplementary Information}
Supplementary Figures~\ref{fig:radar-configuration} and~\ref{fig:signal-processing} detail the acquisition timing and phase-based RHS extraction. The acquisition diagram distinguishes fast-time sampling within each chirp from slow-time observations across frames. The processing diagram connects the complex signal, the selected range-cell phase, the filtered RHS and the reviewed quality intervals.

Reconstruction uses the complex RX3 signal with 512 fast-time samples per observation, the final selected range cell, phase unwrapping, the 79-GHz amplitude conversion constant, forward--backward Butterworth filtering and Savitzky--Golay smoothing. Spectral analysis preserves the original time axis, applies the reviewed quality intervals and excludes one second from each recording edge. Only windows fully contained in retained intervals contribute to the spectrum.

\renewcommand{\thefigure}{S\arabic{figure}}
\renewcommand{\theHfigure}{supplementary.\arabic{figure}}
\setcounter{figure}{0}
\begin{figure}[htbp]
\centering
\includegraphics[width=\textwidth]{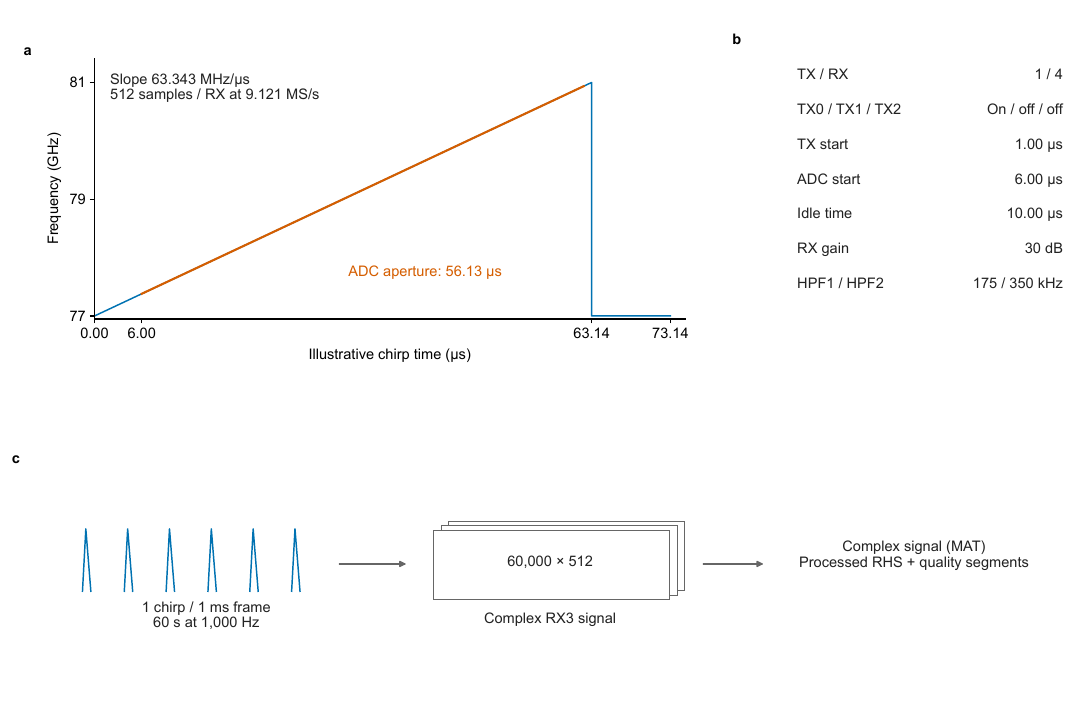}
\caption{RF timing, sampled aperture and raw I/Q representation. a, The 77-GHz-start FMCW ramp, 63.343-MHz/$\mu$s slope, 6-$\mu$s ADC delay and 512 samples at 9.121 MS/s distinguish the sampled aperture from the full ramp. b, The confirmed 1TX4RX configuration enables TX0 and records four receive channels. c, One chirp per 1-ms frame yields 60,000 temporal observations over 60 s; the evaluated receive-channel representation contains 60,000 by 512 complex I/Q samples.}
\label{fig:radar-configuration}
\end{figure}

\begin{figure}[htbp]
\centering
\includegraphics[width=\textwidth]{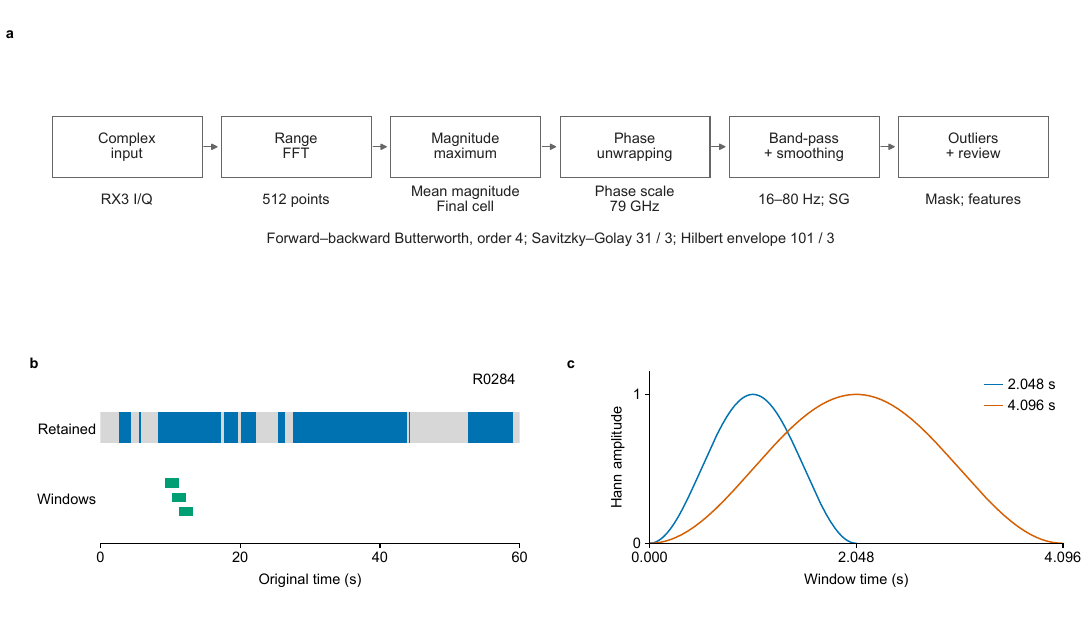}
\caption{Phase-derived radar heart sounds and quality-aware spectral sampling. a, Complex RX3 input, fast-time centring and range FFT, maximum-mean-magnitude target-cell selection, final-cell phase extraction with the 79-GHz conversion constant, forward--backward 16--80-Hz filtering and Savitzky--Golay smoothing, followed by outlier removal and manual quality inspection. Reconstruction uses the final selected cell. The waveform window is 31 points; the illustrated Hilbert envelope uses 101 points. b, Final reviewed intervals in the selected recording with the shortest retained exposure. Blue denotes retained intervals and gray exclusions. The first three complete 2.048-s analysis windows appear in green on separate tracks at their original positions; spectral eligibility additionally excludes the first and last second. No fragments are concatenated. c, Periodic Hann windows of 2,048 and 4,096 samples at 1,000 Hz, evaluated with 50\% overlap. The windows specify the spectral-analysis weighting functions.}
\label{fig:signal-processing}
\end{figure}
}{}
\end{document}